\documentclass[conference]{IEEEtran}
\IEEEoverridecommandlockouts
\usepackage{cite}
\usepackage{amsmath,amssymb,amsfonts}
\usepackage{algorithmic}
\usepackage{graphicx}
\usepackage{textcomp}
\usepackage{xcolor}

\usepackage{verbatim}
\usepackage{booktabs}
\usepackage{multirow}
\usepackage{colortbl}
\usepackage{graphicx}
\usepackage{subfigure}
\usepackage{makecell}

\def\BibTeX{{\rm B\kern-.05em{\sc i\kern-.025em b}\kern-.08em
    T\kern-.1667em\lower.7ex\hbox{E}\kern-.125emX}}
\begin{document}

\title{A Deep Graph Wavelet Convolutional Neural Network for Semi-supervised Node Classification}

% \author{Paper N-0442}
% \begin{comment}

\author{
\IEEEauthorblockN{Jingyi Wang\IEEEauthorrefmark{1}, Zhidong Deng\IEEEauthorrefmark{2}}
Institute for Artificial Intelligence at Tsinghua University (THUAI)\\
State Key Laboratory of Intelligent Technology and Systems\\
Beijing National Research Center for Information Science and Technology (BNRist)\\
Center for Intelligent Connected Vehicles and Transportation\\
Department of Computer Science, Tsinghua University, Beijing 100084, China\\
\IEEEauthorblockA{
\IEEEauthorrefmark{1}wang-jy20@mails.tsinghua.edu.cn}
\IEEEauthorblockA{
\IEEEauthorrefmark{2}michael@mail.tsinghua.edu.cn}
}
% \end{comment}

\maketitle

\begin{abstract}
Graph convolutional neural network provides good solutions for node classification and other tasks with non-Euclidean data. 
There are several graph convolutional models that attempt to develop deep networks but do not cause serious over-smoothing at the same time.
Considering that the wavelet transform generally has a stronger ability to extract useful information than the Fourier transform, we propose a new deep graph wavelet convolutional network (DeepGWC) for semi-supervised node classification tasks. 
Based on the optimized static filtering matrix parameters of vanilla graph wavelet neural networks and the combination of Fourier bases and wavelet ones, DeepGWC is constructed together with the reuse of residual connection and identity mappings in network architectures.
Extensive experiments on three benchmark datasets including Cora, Citeseer, and Pubmed are conducted. 
The experimental results demonstrate that our DeepGWC outperforms existing graph deep models with the help of additional wavelet bases and achieves new state-of-the-art performances eventually.
\end{abstract}
\begin{IEEEkeywords}
graph convolutional neural network, wavelet transform, filtering matrix, network architecture
\end{IEEEkeywords}

\section{Introduction}
Convolutional neural networks (CNNs) achieve outstanding performance in a wide range of tasks that are based on Euclidean data such as computer vision\cite{cnn_cv} and recommender systems\cite{recommend_1}\cite{recommend_2}. 
However, graph-structured non-Euclidean data is also very common in real life. Many research areas like literature citation networks and knowledge graphs have the graph data structure. In this case, graph convolutional networks (GCNs) often have an outstanding exhibition, which are successfully applied in social analysis\cite{social_1}\cite{social_2}, citation network\cite{citation_1}\cite{citation_2}\cite{citation_3}, transport forecasting\cite{transport_1}\cite{transport_2}, and other promising fields. 
For example, in a literature citation network, articles are usually represented as nodes and citation relationships as edges among nodes. When addressing the challenge of semi-supervised node classifications on such non-Euclidean graph data, classical GCNs first extract and aggregate features of articles using the graph Fourier transform and the convolution theorem, and then classify unlabeled articles according to output features of graph convolutions. 

GCN and graph wavelet neural network (GWNN)\cite{gwnn} have their own characteristics. 
In fact, GCN-based methods that employ the Fourier transform as their theoretical basis generally have high computational costs caused by eigendecomposition and frequent multiplications between dense matrices. ChebyNet attempts to reduce the computational burden via polynomial approximation but the locality of convolution is still not guaranteed. 
GWNN tries to address the limitations of GCN by using graph wavelet transform as the theoretical basis instead of graph Fourier transform. Graph wavelet transform has good sparsity in both spatial domain and spectral domain. Besides, wavelets in GWNN are highly localized in the vertex domain. Such localized property makes a great contribution to the performance of GWNN\cite{gwnn}. However, GWNN generally uses independent diagonal filtering matrices for every graph convolution layer. As a result, the memory occupied by the network would increase heavily when stacking more layers in GWNN. To avoid exceeding memory, the number of layers in GWNN models on a large dataset is quite limited. Therefore, GWNN fails to extract deep and high-level features of nodes.

The lately published GCN models suffer from so-called over-smoothing when more layers are stacked. In other words, after repeatedly applying Laplacian smoothing, the features of nodes within each connected component of the graph would converge to the same values\cite{citation_1}. There are several models that expand shallow graph convolution methods in order to get better results. For example, APPNP\cite{appnp} and GDC\cite{gdc} relieve the over-smoothing problem by using the Personalized PageRank matrix. Moreover, GCNII\cite{gcnii} makes use of residual skills and thus surpasses the performance of the previous trials, although only the graph Fourier transform is still utilized as its bases.

In this paper, we propose a new deep graph wavelet convolutional network called DeepGWC that improves the deep GCNII model.
First, the diagonal filtering matrix in graph wavelet convolution of vanilla GWNN models is modified. We no longer set separate learnable convolution kernels that are initialized randomly for each convolution layer. By contrast, we specify the filtering matrices of all layers with selected static elements, i.e., the filtering matrix is not updated during training. As a result, the memory required by the model is greatly reduced. 
Second, we make use of graph wavelet transform and attach the optimized wavelet bases to previous graph Fourier bases of GCNs to enhance the feature extraction capability. Considering the high efficiency and high sparsity of graph wavelet transform\cite{gwnn}, the import of wavelet bases does not bring too much computational complexity. 
Following GCNII, we leverage residual connection and identity mappings  to deepen the above graph wavelet convolutional network with the optimized filtering matrix.
Extensive experiments on semi-supervised node classification are finished. On datasets Cora, Citeseer, and Pubmed, our experimental results demonstrate that the new DeepGWC model improves GCNII. DeepGWC is able to extract useful features and effectively relieve the over-smoothing problem at the same time. This allows the proposed method to outperform the baselines with 2, 4, 8, 16, 32, and 64 layers and achieve new state-of-the-art results.

The contributions of our work are summarized as follows.
\begin{enumerate}
\item We optimize the diagonal filtering matrices in graph wavelet convolution of vanilla GWNNs by specifying them with selected static elements that are not updated during training instead of  learnable parameters. In this way, the memory requirement caused by increasing parameters when stacking more graph wavelet convolutional layers is solved.
\item We attach wavelet bases to the previous graph Fourier bases of classical GCNs as new bases of our DeepGWC model to improve the capability of extracting features of GCNII. With the reuse of residual connections and identity mappings of GCNII, DeepGWC becomes a deep graph wavelet convolutional network with the optimized filtering matrix.
\item We perform experiments on three benchmark datasets, i.e. Cora, Citeseer, and Pubmed. Our DeepGWC model achieves state-of-the-art results in the semi-supervised node classification task.
\end{enumerate}

\section{Related Work}
\subsection{Graph Convolutional Neural Network}
CNN achieves great performance on computer vision\cite{cnn_cv}, natural language processing\cite{cnn_nlp}, and other fields\cite{casfe}\cite{casfe_1}\cite{casfe_11}\cite{casfe_21}. Its success motivates researchers to define convolution operators on graphs. In this way, CNN can be generalized to graphs like social networks and citation networks. Related methods are classified into spatial methods and spectral ones.

Spatial methods perform extraction of spatial features for topological graphs based on similar ideas as weighted summation in CNN. Using different sampling methods for neighborhood nodes, such methods calculate the weighted sum of features of sampled nodes and then update node features. For example, MoNet\cite{monet} computes weighted neighborhood node features instead of using average values. GraphSAGE\cite{graphsage} presents an inductive framework that leverages node feature information to efficiently give rise to node embeddings for previously unseen data. GraphsGAN\cite{graphsgan} generates fake samples to improve performance on graph-based semi-supervised learning. 

Spectral methods carry out the convolution on graphs with the help of spectral theory. They perform convolution operations on topological graphs in accordance with the theory of spectral graph. In fact, the concept of graph Fourier transform is derived from graph signal processing. According to graph Fourier transform, graph convolution is defined. With the rapid development of deep learning methods, GCN in spectral methods also appear. Spectral CNN\cite{spectral_cnn} first implements CNN on graphs using graph Fourier transform. GCN\cite{gcn} motivates the choice of convolutional architecture via a localized first-order approximation of spectral graph convolutions. 

\subsection{Graph Wavelet Neural Networks}
Wavelet transform develops the idea of localization of short-time Fourier transform (STFT) and overcomes the weakness that the window size does not change with frequency at the same time. The wavelet transform is a signal encoding algorithm that is sparser than Fourier transform and has a very strong capability of information expression. 

D.K. Hammond proposes a method for constructing wavelet transforms of functions defined on the vertices of an arbitrary finite weighted graph\cite{wavelet_1}. GraphWave\cite{graphwave} learns a multidimensional structural embedding for each node based on the diffusion of a spectral graph wavelet centered at the node. It provides mathematical guarantees on the optimality of learned structural embeddings. Furthermore, GWNN\cite{gwnn}  addresses the shortcomings of previous spectral graph CNN methods with graph Fourier transform. It takes graph wavelets as a set of bases instead of eigenvectors of graph Laplacian. GWNN redefines graph convolution based on graph wavelets, achieving high efficiency and high sparsity. Besides, GWNN also detaches the feature transformation from graph convolution and thus yields better results.

\subsection{Deep Graph Convolution Models}
It has been proved that features of nodes in a graph would converge to the same values when repeatedly applying graph convolution operations on graphs\cite{citation_1}. In order to relieve such an over-smoothing problem and further enhance the feature extraction capability, several modified models for citation networks have been proposed. For example, JKNet\cite{jknet} makes use of the jump connection and the adaptive aggregation mechanism to adapt to local neighborhood properties and tasks. IncepGCN takes advantage of the inception network and could be optimized by Dropedge\cite{dropedge}. In particular, GCNII\cite{gcnii} is viewed as an effective deep graph convolutional network, which is analyzed that a $K$-layer GCNII model can express a polynomial spectral filtering of order $K$ with arbitrary coefficients. Though various efforts have been made to improve the performance of deep graph convolution models, it is still shallow models that usually obtain the best result.

\section{Methods}
Considering that graph wavelets have a very strong ability to express information, this paper proposes a new deep graph wavelet convolutional network (DeepGWC) to improve the deep graph convolution models\cite{gcnii}. Our DeepGWC model not only achieves the new state-of-the-art performance, but also could  be converted into vanilla GCN, GWNN, and other shallow models by simply adjusting hyperparameters for different application scenarios.

\subsection{Preliminary}
Let $G=(V, E)$ be a simple and connected undirected graph, where $V$ represents the set of nodes with $|V|=n$ and $E$ denotes the set of edges among nodes in $V$ with $|E|=m$. Assume $\textbf{A}$ to express the adjacency matrix of $G$.  $\textbf{D}$ indicates the diagonal degree matrix with $\textbf{D}_{i,i}=\sum_j{\textbf{A}_{i,j}}$. Let $\textbf{L}=\textbf{I}_{n}-\textbf{D}^{-\frac{1}{2}}\textbf{A}\textbf{D}^{-\frac{1}{2}}$ stand for the normalized Laplacian matrix, where $\textbf{I}_n$ is the identity matrix with dimensions $n\times n$.  Obviously, $\textbf{L}$ is a symmetric positive semidefinite matrix with eigendecomposition $\textbf{L}=\textbf{U}\Lambda\textbf{U}^T$, where $\textbf{U}\in \mathbb{R}^{n\times n}$ denotes the complete set of orthonormal eigenvectors and $\Lambda$ is a diagonal matrix of  real, non-negative eigenvalues. 

We use $\tilde{G}=(V, \tilde{E})$ to indicate the cyclic graph of $G$, i.e.,  the graph with a cycle attached to each node in $G$. Correspondingly, the adjacency matrix of $\tilde{G}$ is $\tilde{\textbf{A}}=\textbf{A}+\textbf{I}_n$ and the diagonal degree matrix is  $\tilde{\textbf{D}}=\textbf{D}+\textbf{I}_n$. Let $\tilde{\textbf{L}}=\textbf{I}_{n}-\tilde{\textbf{D}}^{-\frac{1}{2}}\tilde{\textbf{A}}\tilde{\textbf{D}}^{-\frac{1}{2}}$ expresses the normalized Laplacian matrix of the cyclic graph $\tilde{G}$.

\subsection{GCN}
\subsubsection{Graph Fourier Transform}
Graph Fourier transform takes the eigenvectors of the normalized Laplacian matrix, i.e., $\textbf{U}$ as a set of bases. $\hat{\textbf{x}}=\textbf{U}^{\rm T}\textbf{x}$ performs graph Fourier transform on a signal $\textbf{x}\in \mathbb{R}^n$ of graph $G$.
\subsubsection{Graph Convolution Operation}
Based on graph Fourier transform, the graph convolution operation with a filter $g_\gamma(\Lambda)=$diag$(\gamma)$ is defined as
\begin{equation}
\label{equ-graphconvolutionopration}
g_\gamma(\textbf{L})\ast \textbf{x}=\textbf{U}g_\gamma(\Lambda)\textbf{U}^{\rm T}\textbf{x}
\end{equation}
 where $\gamma \in  \mathbb{R}^n$ represents the vector of spectral filtering coefficients. ChebyNet\cite{chebynet}  restricts the convolution kernel $g$ to a polynomial expansion, i.e.,
 \begin{equation}
 \label{equ-convolutionkernel}
 g_\theta = \sum_{k=0}^{K} \theta_{k}\Lambda^{k}
 \end{equation}
where $K$ is a hyperparameter to control the size of node neighborhoods. $\theta \in \mathbb{R}^{K+1}$ stands for a vector of polynomial coefficients. Furthermore, with $\textbf{L}=\textbf{U}\Lambda\textbf{U}^{\rm T}$, the graph convolution operation in Eq. \eqref{equ-graphconvolutionopration} can be approximated by, 
\begin{equation}
\label{equ-graphconvolutionoprationapp}
\textbf{U}g_\theta(\Lambda)\textbf{U}^{\rm T}\textbf{x} \approx 
\textbf{U}(\sum_{k=0}^{K} \theta_{k}\Lambda^{k})\textbf{U}^{\rm T}\textbf{x}=
(\sum_{k=0}^{K} \theta_{k}\textbf{L}^{k})\textbf{x}.
\end{equation}
With $K=1, \theta_0 =2\theta$, and $\theta_1=-\theta$, and the renormalization trick\cite{gcn}, we denote $\tilde{\textbf{P}}=\tilde{\textbf{D}}^{-\frac{1}{2}}\tilde{\textbf{A}}\tilde{\textbf{D}}^{-\frac{1}{2}}$ for the $l$-th graph convolution layer. We have
\begin{equation}
\label{equ-graph_convolution_layer}
\textbf{H}^{l+1}=\sigma(\tilde{\textbf{P}}\textbf{H}^l \textbf{W}^l)
\end{equation}
where $\sigma$ expresses the activation function and $\textbf{W}^l \in \mathbb{R}^{p \times q}$ is the parameter matrix for feature transformation. The graph convolution layer in Eq. \eqref{equ-graph_convolution_layer} transforms the input feature matrix of $\textbf{H}^l \in \mathbf{R}^{n\times p}$ into the output feature matrix of $\textbf{H}^{l+1} \in \mathbf{R}^{n\times q}$.

\subsection{Graph Wavelet Convolutional Network}
\subsubsection{Graph Wavelet Transform}
Similar to graph Fourier transform, the graph wavelet transform need a proper set of bases to project graph signal from vertex domain into the spectral domain. Here we indicate the wavelet bases as $\psi_s$, where $s$ is a scaling parameter. $\psi_s$ can be defined by
\begin{equation}
\label{equ-graph_wavelet_bases}
\psi_s=\textbf{U}\textbf{G}_s\textbf{U}^{\rm{T}}
\end{equation}
where $\textbf{G}_s$ is a scaling matrix that has similar effects to $g_\gamma(\Lambda)$ in graph Fourier transform. In this case, $\textbf{G}_s$ can be evaluated by $\textbf{G}_s={\rm diag} (g(s\Lambda))$, where $g$ is an exponential function. We can get $\psi_s^{-1}$ by replacing $s$ with $-s$. By using  graph wavelets as bases, $\hat{\textbf{x}}=\psi_s\textbf{x}$ conducts graph Fourier transform on a signal $\textbf{x}\in \mathbb{R}^n$ of graph $G$.

\subsubsection{Graph Wavelet Convolution Operation}
\label{sub-gwco}
Substituting Fourier transform in graph convolution operation with wavelet transform, we can find the structure of the $l$-th graph wavelet convolution layer below:
\begin{equation}
\label{equ-graph_wavelet_convolution_layer}
\textbf{H}^{l+1}=\sigma(\psi_s \textbf{F} \psi_s^{-1} \textbf{H}^l \textbf{W}^l)
\end{equation}
 where $\textbf{F}$ indicates  the diagonal matrix for graph convolution kernel\cite{gwnn}. In Eq. \eqref{equ-graph_wavelet_convolution_layer}, $\sigma$, $\textbf{W}^l$, $\textbf{H}^l$, and $\textbf{H}^{l+1}$ have the same definitions as those in Eq. \eqref{equ-graph_convolution_layer}.
 
For vanilla GWNN, $\textbf{F}$ is a diagonal filtering matrix learned in the spectral domain and independent for every layer. In experiments done in \cite{gwnn}, a two-layer graph wavelet neural network is just designed for three datasets. In this case, the memory consumption is not too large for such a shallow GWNN. However, when trying to increase the number of layers to 8 or 16, the memory usage would rise rapidly. Actually, the out-of-memory problem is particularly prone to appear in the experiments on the Pubmed dataset. In order to solve such difficulties, we simplify the use of diagonal filtering matrix $\textbf{F}$. Instead of exploiting independent filtering matrices for every layer, $\textbf{F}$ with the same static parameters is employed. By optimizing the elements of diagonal elements of $\textbf{F}$ as a set of static parameters, the wavelet basis can also be obtained statically like the Fourier basis. In this way, we add no parameters to be trained.

\subsection{Network Architecture against Over-Smoothing}
\subsubsection{Residual Connection}
The initial residual connection constructs a link to the initial feature representation $H^0$\cite{gcnii}. We make sure that the final representation still contains a portion of the initial feature no matter how many layers we stack\cite{gcnii}.

\subsubsection{Identity Mapping}
Like the identity mapping in ResNet\cite{resnet}, we append an identity matrix to the weight matrix $W^l$ in the case of $p=q$. Thus a direct pathway is established between the input and the output of a convolutional layer through importing identity mapping\cite{gcnii}.

\subsection{Our DeepGWC Model}
\label{sec-DeepGWC}
In this paper, we build a deep graph wavelet  convolutional network (DeepGWC). On the basis of the above-mentioned, we define the $l$-th layer of DeepGWC as
\begin{equation}
\label{equ-DeepGWC_layer}
\textbf{H}^{l+1}=\sigma(\textbf{H}^{l^\prime}\textbf{W}^{l^\prime})
\end{equation}
where $\sigma$ is the activation function, $\textbf{H}^{l^\prime}$ is the results of graph convolution on $\textbf{H}$, and $\textbf{W}^{l^\prime}$ is the optimized feature transformation matrix. $\textbf{H}^{l^\prime}$ and $\textbf{W}^{l^\prime}$ in Eq. \eqref{equ-DeepGWC_layer} are described as follows:
\begin{equation}
\label{equ-hl'}
\textbf{H}^{l^\prime}=(1-\alpha)\tilde{\textbf{P}}^\prime \textbf{H}^l+\alpha \textbf{H}^0,
\end{equation}
\begin{equation}
\label{equ-wl'}
\textbf{W}^{l^\prime}=\beta_l \textbf{W}^l+(1-\beta_l)\textbf{I}
\end{equation}
where $\alpha$ represents the ratio of the initial residual term $\textbf{H}^0$, $\beta_l$ is the ratio of the original feature transformation matrix $\textbf{W}^l$ of the $l$-th layer, and $0\leq \alpha,\beta \leq 1$.  $\tilde{\textbf{P}}^\prime$ in Eq. \eqref{equ-hl'} is given by \begin{equation}
\label{equ-p'}
\tilde{\textbf{P}}^\prime = \gamma(\psi_s \textbf{F} \psi_s^{-1})+(1-\gamma)(\tilde{\textbf{D}}^{-\frac{1}{2}}\tilde{\textbf{A}}\tilde{\textbf{D}}^{-\frac{1}{2}})
\end{equation}
where $\gamma$ stands for the ratio of the graph wavelet term and $0\leq \gamma\leq 1$. As described in subsection \ref{sub-gwco}, $\textbf{F}$ is optimized with static elements, i.e., the modified $\textbf{F}$ is the product of a static constant $f$ and the identity matrix. In fact, $\gamma$ and $f$ could be combined and the first item of Eq. \eqref{equ-p'} could be written as
\begin{equation}
\label{equ-combinegammaf}
\gamma(\psi_s \textbf{F} \psi_s^{-1})=\gamma f (\psi_s \psi_s^{-1}).
\end{equation}
However, to emphasize that we combine the Fourier bases and the wavelet bases, we still treat $\gamma$ and $f$ as two separate parameters later, i.e., we perform experiments according to Eq. \eqref{equ-p'}. In this way, we could keep consistency with the form of vanilla wavelet bases in Eq. \eqref{equ-graph_wavelet_convolution_layer} to emphasize our modification about the filtering matrix $\textbf{F}$.

There are the three key hyperparameters, i.e., $\alpha$, $\beta$, and $\gamma$ in our DeepGWC model. Through adjusting such three parameters, the  DeepGWC could be reduced to the following basic models like:

\begin{itemize}
\item With $\alpha=0$, $\beta=1$ and $\gamma=0$, Eq. \eqref{equ-DeepGWC_layer} can be simplified to Eq. \eqref{equ-graph_convolution_layer} and our DeepGWC model is equivalent to the vanilla GCN model\cite{gcn}.
\item With $\alpha=0$, $\beta=1$ and $\gamma=1$, Eq. \eqref{equ-DeepGWC_layer} can be simplified to Eq. \eqref{equ-graph_wavelet_convolution_layer} and our DeepGWC model is then reduced to the GWNN model\cite{gwnn}.
\item With $\alpha \neq 0$, $\beta=0$ and $\gamma=0$, the weight matrix $\textbf{W}^l$ is ignored and our DeepGWC model then functions like the APPNP model\cite{appnp}.
\item With $\alpha \neq 0$, $\beta \neq 0$ and $\gamma=0$, the weight matrix $\textbf{W}^l$ is ignored and our DeepGWC model then resembles the GCNII model\cite{gcnii}.
\end{itemize}

\section{Experiments}
We compare the proposed DeepGWC model with competitive baselines, i.e., the existing state-of-the-art graph neural network models on the semi-supervised node classification task.
\subsection{Datasets and Baselines}
\subsubsection{Datasets}
We use three commonly-used citation network datasets, i.e., Cora, Citeseer, and Pubmed\cite{pubmed}. The statistics of datasets are summarized in Table \ref{tab-dataset}. For every dataset, 20 labeled nodes per class are adopted for training in the overall assessment.
\begin{table}[htbp]
  \centering
  \caption{The statistics of three citation network benchmarks}
  \label{tab-dataset}
  \begin{tabular}{cccccc}    
    \toprule
    Dataset & \#Nodes & \#Edges & \#Classes & Features & Label Rate \\
    \midrule
    Cora     & 2,708  & 5,429  & 7       & 1,433     & 5.2\%      \\
    CiteSeer & 3,327  & 4,732  & 6       & 3,703     & 3.6\%      \\
    Pubmed   & 19,717 & 44,338 & 3       & 500      & 0.3\%      \\
    \bottomrule
  \end{tabular} 
\end{table}
\subsubsection{Baselines}
Considering that the vanilla GCN\cite{gcn} is one of the basic model graph convolutional methods, we exploit it as an important baseline. Meanwhile, spectral CNN\cite{spectral_cnn}, ChebyNet\cite{chebynet}, GAT\cite{gat}, APPNP\cite{appnp}, and GWNN\cite{gwnn} are also utilized for performance comparison. In the experiment on the number of stacked layers, we are more interested in comparison with JKNet\cite{jknet}, IncepGCN\cite{dropedge}, and GCNII\cite{gcnii} because these three models are often employed to set up deeper graph convolutional networks. In the experiments on the label rate, we focus on M3S\cite{multistage} and some different training strategies\cite{citation_1} that improve the generalization capability of GCNs on graphs with few labeled nodes.
\subsection{Implementation Details}
There are two hyperparameters in the calculation of graph wavelets in general. $s$ denotes the scaling parameter and $t$ stands for the threshold of $\psi_s$ and $\psi_s^{-1}$. Elements that are less than $t$ are ignored and reset to zero in $\psi_s$ and $\psi_s^{-1}$. 
\begin{comment}
\begin{table}[htbp]
\centering
\caption{Hyperparameters for graph wavelets and densities of wavelets bases}
\label{tab-gwnnst}
\begin{tabular}{cccc}
\toprule
Dataset  & $s$   & $t$    & Density \\
\midrule
Cora     & 1.0 & 1e-4 & 2.81\%   \\
Citeseer & 0.7 & 1e-5 & 1.52\%   \\
Pubmed   & 0.5 & 1e-7 & 5.03\%    \\
\bottomrule
\end{tabular}
\end{table}
\end{comment}
Following the parameter settings of \cite{gwnn}, we select the values of $s$ and $t$.
For Cora, we set $s=1.0$, $t=1e-4$, and the calculated density of wavelet bases is 2.81\%. For Citeseer, we set $s=0.7$, $t=1e-5$, and the density value is 1.52\%. For Pubmed, we set $s=0.5$, $t=1e-7$, and the density value is 5.03\%. We also provide the density values of wavelet bases. The wavelet bases with low density would not give rise to the calculation burden too much.

Note that we employ a linear connection layer before the graph convolution layers of DeepGWC that projects the original feature dimension on the hidden embedding dimension $\textbf{H}^0$. Similarly, there is also a linear connection layer after the graph convolution layers that maps the hidden embedding dimension to the output layer $\textbf{Y}$, which has the same number of neurons as classes delivered in the dataset. Then we perform the logarithm-softmax operation below:
\begin{equation}
\label{equ-logsoftmax}
\textbf{Z}_{i,j}=\log(\frac{e^{\textbf{Y}_{i,j}}}{\sum_k e^{\textbf{Y}_{i,k}}})
\end{equation}
where $\textbf{Z}$ denotes the actual output of the whole DeepGWC model. In the experiments, the negative log likelihood (NLL) loss function is exploited. Moreover, the proportion of nodes classified correctly is used as the metric to evaluate the accuracy of models.  We adopt the Adam optimizer to train our DeepGWC. Following \cite{gcnii}, we set $\beta_l=\log(1+\eta / l)\approx \eta / l$ where $\eta$ is a hyperparameter and $l$ means the $l$-th layer of DeepGWC. Dropout is adopted to avoid the overfitting issue.

\subsection{Overall Assessment}
On Cora, Citeseer, and Pubmed, we follow the parameter settings of the vanilla GCN\cite{gcn} to ensure impartial performance comparison. For every dataset, 20 labeled nodes per class are adopted for training. In order to optimize the parameters of the DeepGWC model, 500 samples are used for validation and 1,000 ones for testing so as to get final results.
\begin{table}[htbp]
  \centering
  \caption{The performance comparison of DeepGWC with existing graph neural network methods}
  \label{tab-results}
  \begin{tabular}{cccc}    
    \toprule
    Methods & Cora & CiteSeer & Pubmed\\
    \midrule
    Spectral CNN\cite{spectral_cnn}     & 73.3\%  & 58.9\%  & 73.9\%     \\
    ChebyNet\cite{chebynet} 			& 81.2\%  & 69.8\%  & 74.4\%     \\
    GCN\cite{gcn}   					& 81.5\%  & 70.3\%  & 79.0\%     \\
    GAT	\cite{gat}					 & 83.1\%  & 70.8\%  & 78.5\%     \\
    APPNP\cite{appnp}				  & 83.3\%  & 71.8\%  & 80.1\%  	\\
    JKNet\cite{jknet}					& 81.1\%  & 69.8\%  & 78.1\% 	  \\
    IncepGCN\cite{dropedge}		  		     & 81.7\%  & 70.2\%  & 77.9\% 	   \\
    GWNN\cite{gwnn}			    & 82.8\%  & 71.7\%  & 79.1\%	  \\
    GCNII\cite{gcnii}				   &85.5\%   & 73.4\%  & 80.3\%		 \\
    \midrule
    \textbf{DeepGWC}          &\textbf{86.4\%}   &\textbf{75.0\%}  &\textbf{81.6\%}    \\
    \bottomrule
  \end{tabular} 
\end{table}
We report the experimental results of the proposed model yielded on three datasets including Cora, CiteSeer, and Pubmed as shown in Table \ref{tab-results}. We quote the experimental results of Spectral CNN\cite{spectral_cnn}, ChebyNet\cite{chebynet}, GCN\cite{gcn}, and GWNN\cite{gwnn} that are already reported in \cite{gwnn}. The results of GAT\cite{gat}, APPNP\cite{appnp}, JKNet\cite{jknet}, and IncepGCN\cite{dropedge} are provided according to \cite{gcnii}. For Cora, CiteSeer, and Pubmed, the depths of JKNet\cite{jknet} are 4, 16, and 32, respectively, while the depths of IncepGCN\cite{dropedge} are 64, 4, and 4, respectively. By contrast, our DeepGWC model has depths of 64, 64, and 32, respectively. The others are all shallow models. 

For the DeepGWC, we utilize the Adam Optimizer with a learning rate of 0.001. The detailed hyperparameters of DeepGWC of Table \ref{tab-results} are given as follows. Let $f$ stand for the element of the filtering matrix $\textbf{F}$, which is searched over (0.4, 0.8, 1.2, 1.6). Suppose that $L$ represents the number of graph wavelet convolution layers in DeepGWC, $d$ represents the dimension of hidden layers. We get the results of Table \ref{tab-results} with the parameters of:
\begin{itemize}
\item For Cora, we set  $L=64$, $d=64$, $\alpha=0.3$, $\eta=0.8$, $\gamma=0.4$, $f=0.4$.
\item For Citeseer, we set  $L=64$,  $d=256$, $\alpha=0.1$, $\eta=0.8$, $\gamma=0.4$, $f=0.4$. 
\item For Pubmed, we set  $L=32$,  $d=512$, $\alpha=0.1$, $\eta=0.4$, $\gamma=0.4$, $f=0.6$. 
\end{itemize}
As for the parameters above, larger datasets require higher hidden feature dimension $d$ to express information as completely as possible. We follow the setting of $\alpha$ and $\eta$ of GCNII\cite{gcnii} basically for the three datasets. The proportion of wavelet bases in the combined bases is reflected in the value of $\gamma \cdot f$, which is 0.16 for Cora and Citeseer, and 0.24 for Pubmed. The proportion of wavelet bases for Pubmed is higher than that for Cora and Citeseer because there are more nodes in Pubmed than Cora and Citeseer obviously. The locality of wavelet bases is important in the case of a large number of nodes.

\begin{figure*}[htbp]
    \centering
\includegraphics[width=16cm]{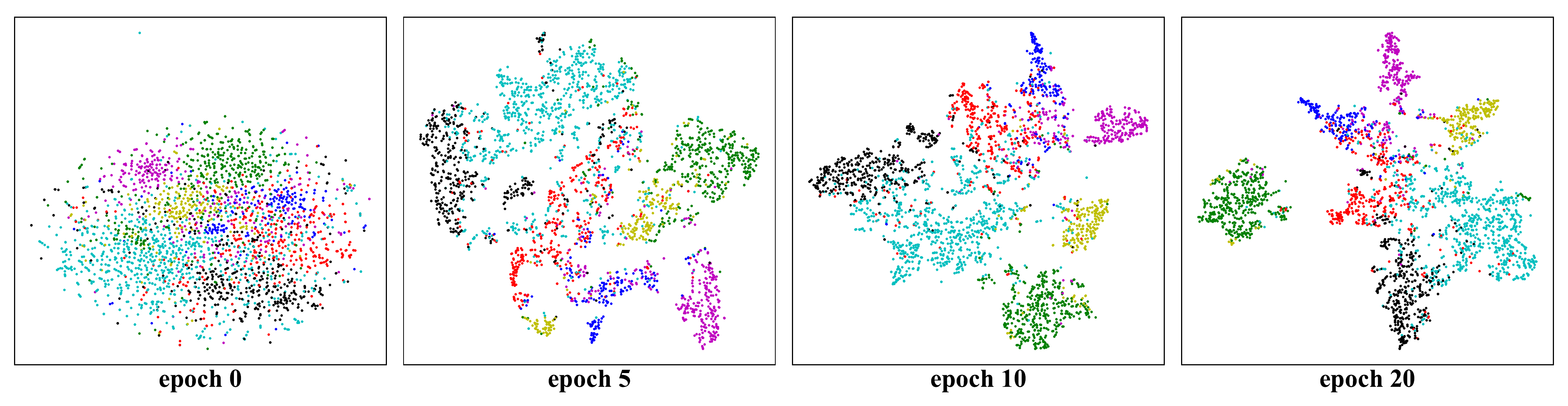}
    \caption{The visualization of learned features of the Cora dataset during training DeepGWC.}
    \label{fig-tsne}
\end{figure*}

Analyzing Table \ref{tab-results}, the DeepGWC improves GCNII\cite{gcnii} and achieves the best results among all the shallow and deep graph models on the three benchmark datasets. The average accuracies on the three datasets of DeepGWC is higher than that of the other models by at least 1.3\%. We believe that the adaptability of our DeepGWC plays an important role in obtaining the best results for semi-supervised node classification problems. DeepGWC outperforms GCNII on every dataset, which indicates the excellence of wavelet bases. In addition, for DeepGWC with 8 graph wavelet convolution layers, we utilize the t-SNE algorithm\cite{tsne} to visualize the learned embeddings during training on the Cora dataset. As shown in Fig.\ref{fig-tsne}, different colors represent different classes of articles. The difference between features of different classes become more obvious as the training proceeds.
 
\subsection{Deeper Network}
We further investigate the influence of network depths on  model performance. Table \ref{tab-results_layers} shows the classification accuracy vs. the network depth. For every dataset, the best accuracy with a fixed number of layers is marked in bold. For specific datasets and methods, the cell with a grey background in each row gives the best result in that row, i.e., the best number of layers. We report the classification results of GCN\cite{gcn}, JKNet\cite{jknet}, IncepGCN\cite{dropedge}, GWNN\cite{gwnn}, GCNII\cite{gcnii}, and our DeepGWC model with 2, 4, 8, 16, 32, and 64 layers, where Dropedge\cite{dropedge} is further equipped with GCN\cite{gcn}, JKNet\cite{jknet}, and IncepGCN\cite{dropedge}, respectively. 

\begin{table}[htbp]
\setlength\tabcolsep{1.5pt}
\centering
\caption{The results of Node Classifications with different depths}
\label{tab-results_layers}
\begin{tabular}{c|c|cccccc}
\toprule
{\multirow{2}{*}{Dataset}} & {\multirow{2}{*}{Method}} & \multicolumn{6}{c}{\#Layers} \\
& & 2  & 4  & 8 & 16 & 32 & 64\\
\midrule
\multirow{9}{*}{\makecell[c]{Cora \\ \cite{pubmed}}}
&{GCN\cite{gcn}}&\cellcolor [gray]{.8}81.1\%&80.4\%&69.5\%&64.9\%&60.3\%&28.7\%\\
&{GCN(drop)\cite{gcn}\cite{dropedge}}&\cellcolor [gray]{.8}82.8\%&82.0\%&75.8\%&75.7\%&62.5\%&49.5\%\\
&{JKNet\cite{jknet}}&-&80.2\%&80.7\%&80.2\%&\cellcolor [gray]{.8}81.1\%&71.5\%\\
&{JKNet(drop)\cite{jknet}\cite{dropedge}}&-&\cellcolor [gray]{.8}83.3\%&82.6\%&83.0\%&82.5\%&83.2\%\\
&{IncepGCN\cite{dropedge}}&-&77.6\%&76.5\%&81.7\%&\cellcolor [gray]{.8}81.7\%&80.0\%\\
&{IncepGCN(drop)\cite{dropedge}}&-&82.9\%&82.5\%&83.1\%&83.1\%&\cellcolor [gray]{.8}83.5\%\\
&{GWNN\cite{gwnn}}&\cellcolor [gray]{.8}82.8\%&78.2\%&50.5\%&33.7\%&33.4\%&31.9\%\\
&{GCNII\cite{gcnii}} &82.2\%&82.6\%&84.2\%&84.6\%&85.4\%&8\cellcolor [gray]{.8}5.5\%\\
\specialrule{0em}{0.5pt}{0.5pt}
\cline{2-8}
\specialrule{0em}{0.5pt}{0.5pt}
&{\textbf{DeepGWC}}&\textbf{84.4}\%&\textbf{85.6\%}&\textbf{85.5\%}&\textbf{86.3\%}&\textbf{86.2\%}&\cellcolor [gray]{.8}\textbf{86.4\%}\\

\midrule
\multirow{9}{*}{\makecell[c]{Citeseer \\ \cite{pubmed}}}
&{GCN\cite{gcn}}&\cellcolor [gray]{.8}70.8\%&67.6\%&30.2\%&18.3\%&25.0\%&20.0\%\\
&{GCN(drop)\cite{gcn}\cite{dropedge}}&\cellcolor [gray]{.8}72.3\%&70.6\%&61.4\%&57.2\%&41.6\%&34.4\%\\
&{JKNet\cite{jknet}}&-&68.7\%&67.7\%&\cellcolor [gray]{.8}69.8\%&68.2\%&63.4\%\\
&{JKNet(drop)\cite{jknet}\cite{dropedge}}&-&72.6\%&71.8\%&\cellcolor [gray]{.8}72.6\%&70.8\%&72.2\%\\
&{IncepGCN\cite{dropedge}}&-&69.3\%&68.4\%&\cellcolor [gray]{.8}70.2\%&68.0\%&67.5\%\\
&{IncepGCN(drop)\cite{dropedge}}&-&\cellcolor [gray]{.8}72.7\%&71.4\%&72.5\%&72.6\%&71.0\%\\
&{GWNN\cite{gwnn}}&\cellcolor [gray]{.8}71.7\%&64.0\%&45.6\%&20.2\%&21.2\%&15.5\%\\
&{GCNII\cite{gcnii}} &68.2\%&68.9\%&70.6\%&72.9\%&\cellcolor [gray]{.8}73.4\%&73.4\%\\
\specialrule{0em}{0.5pt}{0.5pt}
\cline{2-8}
\specialrule{0em}{0.5pt}{0.5pt}
&{\textbf{DeepGWC}}&\textbf{72.6}\%&\textbf{73.6\%}&\textbf{73.3\%}&\textbf{74.4\%}&\cellcolor [gray]{.8}\textbf{75.0\%}&\textbf{74.9\%}\\

\midrule
\multirow{9}{*}{\makecell[c]{Pubmed \\ \cite{pubmed}}}
&{GCN\cite{gcn}}&\cellcolor [gray]{.8}79.0\%&76.5\%&61.2\%&40.9\%&22.4\%&35.3\% \\
&{GCN(drop)\cite{gcn}\cite{dropedge}}&\cellcolor [gray]{.8}79.6\%&79.4\%&78.1\%&78.5\%&77.0\%
&61.5\% \\
&{JKNet\cite{jknet}}&-&78.0\%&\cellcolor [gray]{.8}78.1\%&72.6\%&72.4\%&74.5\% \\
&{JKNet(drop)\cite{jknet}\cite{dropedge}}&-&78.7\%&78.7\%&79.1\%&\cellcolor [gray]{.8}79.2\%&78.9\% \\
&{IncepGCN\cite{dropedge}} &-&77.7\%&\cellcolor [gray]{.8}77.9\%&74.9\%&OOM&OOM\\
&{IncepGCN(drop)\cite{dropedge}}&-&\cellcolor [gray]{.8}79.5\%&78.6\%&79.0\%&OOM&OOM \\
&{GWNN\cite{gwnn}}&\cellcolor [gray]{.8}79.1\%&76.4\%&OOM&OOM&OOM&OOM \\
&{GCNII\cite{gcnii}}&78.2\%&78.8\%&79.3\%&\cellcolor [gray]{.8}80.2\%&79.8\%&79.7\% \\
% \cmidrule(l){2-8}
\specialrule{0em}{0.5pt}{0.5pt}
\cline{2-8}
\specialrule{0em}{0.5pt}{0.5pt}
&{\textbf{DeepGWC}}&\textbf{80.4}\%&\textbf{81.0\%}&\textbf{81.1\%}&\textbf{81.0\%}&\cellcolor [gray]{.8}\textbf{81.6\%}&\textbf{80.7\%}\\
% \hline
\bottomrule
\end{tabular}
\end{table}

It is easy to see from Table \ref{tab-results_layers} that our DeepGWC model improves GCNII\cite{gcnii} and yields the best results with all numbers of layers. The accuracy of the DeepGWC model becomes better roughly as the number of layers increases. GCN\cite{gcn} and dropped GCN\cite{dropedge} do have good results with 2 layers, but their performance gets worse rapidly when stacking much more layers. Meanwhile, JKNet\cite{jknet} and IncepGCN\cite{dropedge} reach their best results with higher layers, but the accuracy is still not guaranteed when the number of layers exceeds 32 layers. GWNN\cite{gwnn} suffers more from the over-smoothing problem than the vanilla GCN\cite{gcn}. Besides, there exists a serious memory issue on Pubmed when using GWNN\cite{gwnn} with more than 4 layers. Among the baselines, the performance of GCNII\cite{gcnii} is exceptionally good because of its success in reliving the over-smoothing problem. The proposed DeepGWC improves GCNII with wavelet bases further.
\begin{figure*}
    \centering
    \subfigure[Cora]{
    \includegraphics[width=5cm]{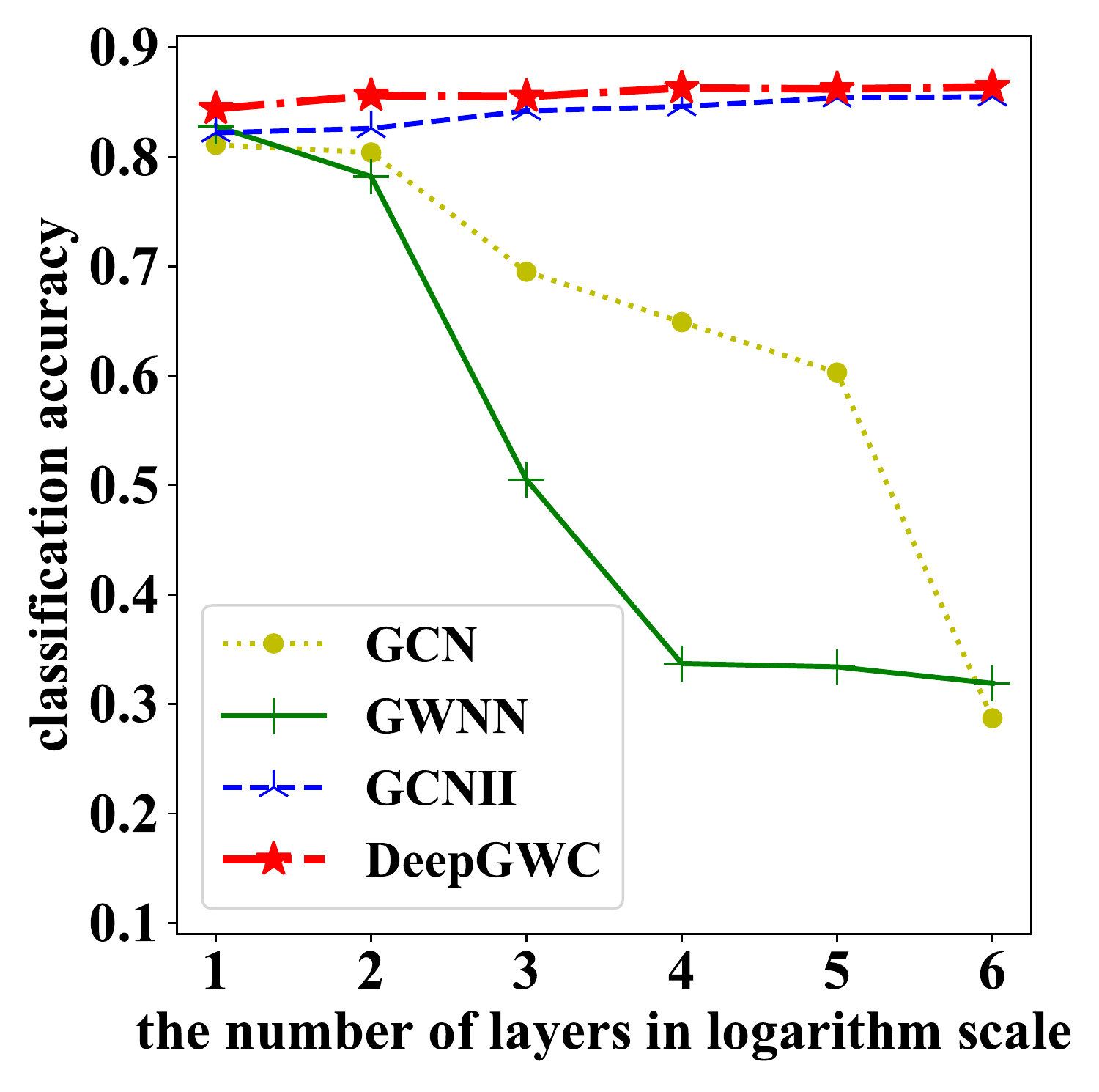}
    }
    \subfigure[Citeseer]{
    \includegraphics[width=5cm]{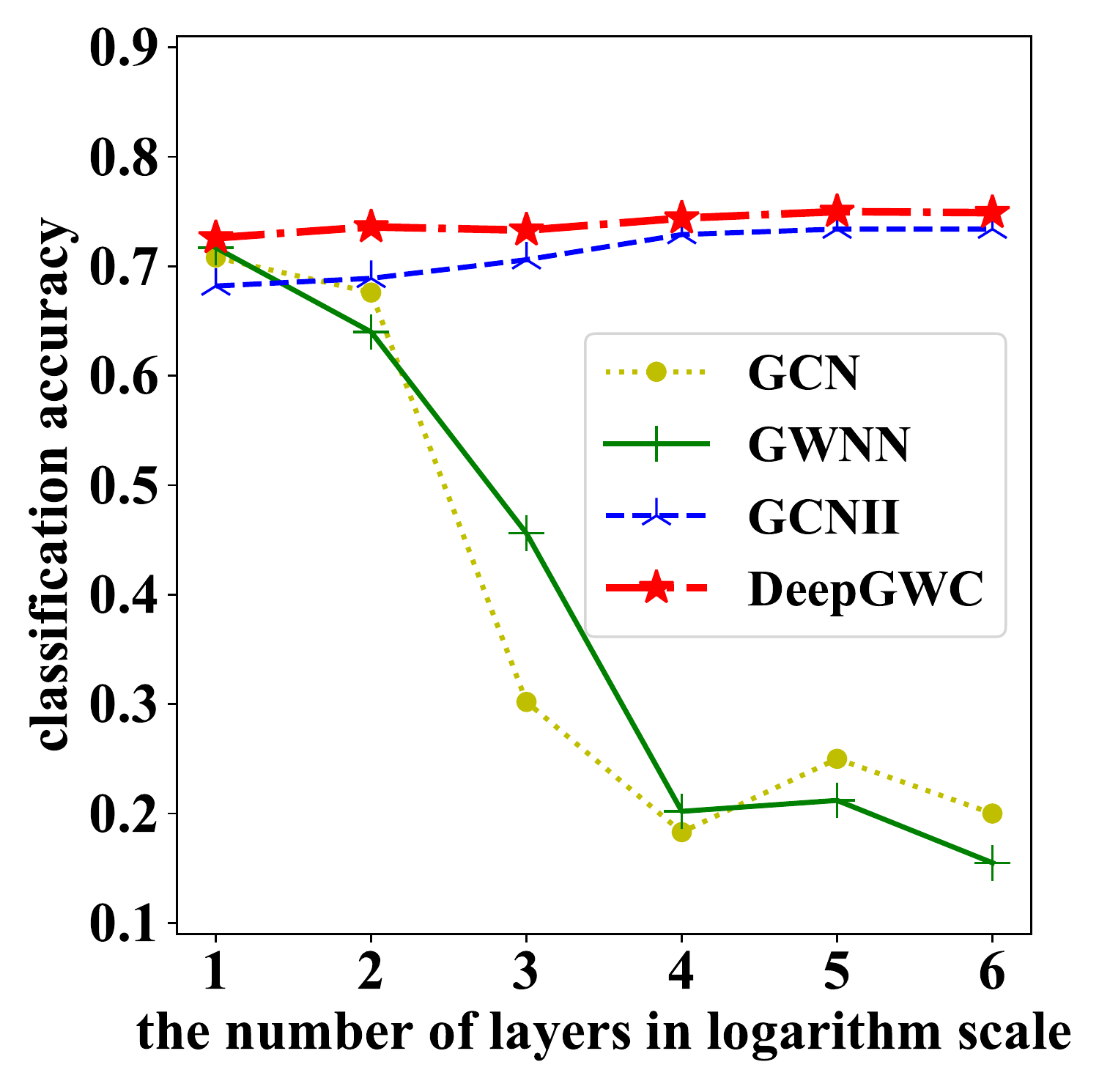}
    }
    \subfigure[Pubmed]{
    \includegraphics[width=5cm]{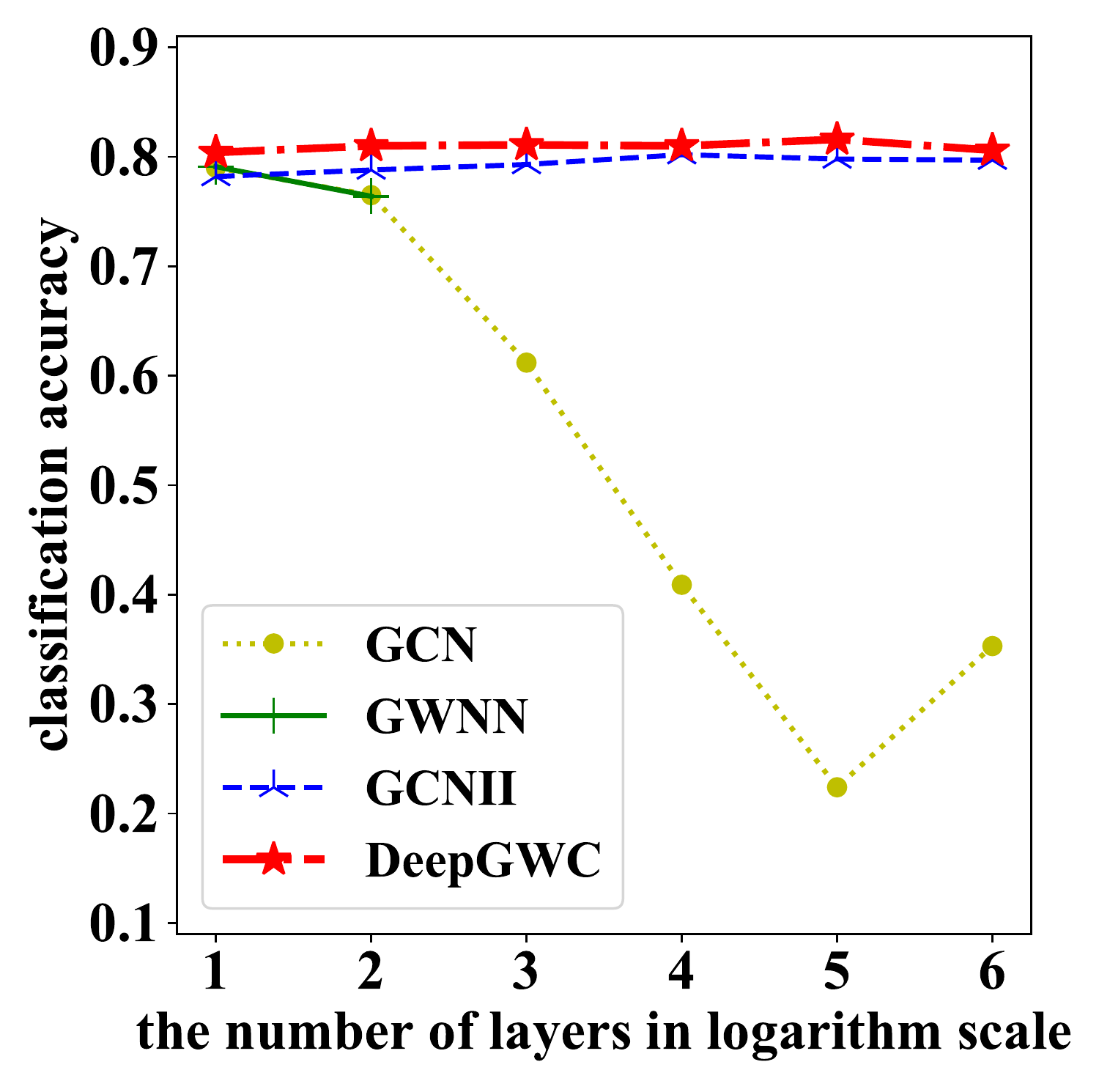}
    }
    \caption{Performance comparisons of DeepGWC with GCN, GWNN, and GCNII as ablation study.}
    \label{fig-ablation}
\end{figure*}

We accomplish the comparison of DeepGWC with all the converted models  GCN\cite{gcn}, GWNN\cite{gwnn}, and GCNII\cite{gcnii} models as described in Subsection \ref{sec-DeepGWC} and accordingly complete the ablation study. Fig.\ref{fig-ablation} shows the classification accuracy vs. the number of layers of our DeepGWC model and the three relevant baselines which motivate our study. It is readily observed from Fig. \ref{fig-ablation} that the performance of both GCN\cite{gcn} and GWNN\cite{gwnn} draws rapidly with the increase of the number of layers, while both GCNII\cite{gcnii} and DeepGWC steadily get better performance when stacking more layers. As a result of the addition of graph wavelet bases, our DeepGWC model achieves better performance than GCNII\cite{gcnii} in the semi-supervised node classification task.

\subsection{Fewer Training Samples}
We complete experiments on training with less labeled nodes. MultiStage\cite{multistage} and M3S\cite{multistage} are two training methods  that enable GCNs to learn graph embeddings with few supervised signals. Following the experiments of M3S\cite{multistage}, we use Label Propagation (LP)\cite{lp} and GCN\cite{gcn} as baselines. For GCN\cite{gcn}, we adopt four training strategies, i.e., Co-training, Self-training, Union, and Intersection additionally\cite{citation_1}. For Cora and Citeseer, we train our DeepGWC model and the baselines with the label rates of 0.5\%, 1\%, 2\%, 3\%, and 4\%. For Pubmed, we train the models with the label rates of 0.03\%, 0.05\%, and 0.1\% because Pubmed has more nodes.  500 samples are employed for validation and 1,000 ones for testing so as to get final results. We exploit the same number of layers and hyperparameters as that in Table \ref{tab-results} for GWNN\cite{gwnn}, GCNII\cite{gcnii}, and DeepGWC. We only adjust the label rates.

Table \ref{tab-lesscora}, Table \ref{tab-lessciteseer}, and Table \ref{tab-lesspubmed} list the classification results with fewer training samples. Our DeepGWC model obtains the best performance on most of the experiments with lower label rates. The accuracy of DeepGWC is 5.88\% higher than that of M3S\cite{multistage} on average, which  focuses on improving the generalization capability of GCNs on graphs with few labeled nodes. 

\begin{table}[htbp]
\caption{The classification results with less training samples on Cora}
\label{tab-lesscora}
\centering
\begin{tabular}{c|ccccc}
\toprule
\multirow{2}{*}{Method} & \multicolumn{5}{c}{Label Rate}             \\
                        & 0.5\%  & 1\%    & 2\%    & 3\%    & 4\%    \\
\midrule
LP\cite{lp}                      & 57.6\% & 61.0\% & 63.5\% & 64.3\% & 65.7\% \\
GCN\cite{gcn}                     & 50.6\% & 58.4\% & 70.0\% & 75.7\% & 76.5\% \\
Co-training\cite{citation_1}             & 53.9\% & 57.0\% & 69.7\% & 74.8\% & 75.6\% \\
Self-training\cite{citation_1}           & 56.8\% & 60.4\% & 71.7\% & 76.8\% & 77.7\% \\
Union\cite{citation_1}                   & 55.3\% & 60.0\% & 71.7\% & 77.0\% & 77.5\% \\
Intersection\cite{citation_1}            & 50.6\% & 60.4\% & 70.0\% & 74.6\% & 76.0\% \\
MultiStage\cite{multistage}              & 61.1\% & 63.7\% & 74.4\% & 76.1\% & 72.2\% \\
M3S\cite{multistage}                     & 61.5\% & 67.2\% & 75.6\% & 77.8\% & 78.0\% \\
GWNN\cite{gwnn}                    & 60.8\% & 70.7\% & 77.1\% & 77.9\% & 80.1\% \\
GCNII\cite{gcnii}& 62.3\% & 69.4\% & 77.6\% & 81.7\% & 84.2\% \\
\midrule
\textbf{DeepGWC}                   & \textbf{66.0\%} & \textbf{73.0\%} & \textbf{80.3\%} & \textbf{82.0\% }& \textbf{84.7\%}\\
\bottomrule
\end{tabular}
\end{table}

\begin{table}[htbp]
\caption{The classification results with less training samples on Citeseer}
\label{tab-lessciteseer}
\centering
\begin{tabular}{c|ccccc}
\toprule
\multirow{2}{*}{Method} & \multicolumn{5}{c}{Label Rate}\\
& 0.5\%  & 1\%    & 2\%    & 3\%    & 4\%    \\
\midrule
LP\cite{lp}& 37.7\% & 41.6\% & 41.9\% & 44.4\% & 44.8\% \\
GCN\cite{gcn}& 44.8\% & 54.7\% & 61.2\% & 67.0\% & 69.0\% \\
Co-training\cite{citation_1}& 42.0\% & 50.0\% & 58.3\% & 64.7\% & 65.3\% \\
Self-training\cite{citation_1}& 51.4\% & 57.1\% & 64.1\% & 67.8\% & 68.8\% \\
Union\cite{citation_1}& 48.5\% & 52.6\% & 61.8\% & 66.4\% & 66.7\% \\
Intersection\cite{citation_1}& 51.3\% & 61.1\% & 63.0\% & 69.5\% & 70.0\% \\
MultiStage\cite{multistage}& 53.0\% & 57.8\% & 63.8\% & 68.0\% & 69.0\% \\
M3S\cite{multistage}& 56.1\% & 62.1\% & 66.4\% & 70.3\% & 70.5\% \\
GWNN\cite{gwnn}& 53.3\% & 55.7\% & 65.7\% & 69.4\% & 71.3\% \\
GCNII\cite{gcnii}& 52.6\% & 58.7\% & 70.0\% & 72.2\% & \textbf{73.9\%} \\
\midrule
\textbf{DeepGWC}&\textbf{57.9\%} &\textbf{62.5\%} &\textbf{70.2\% }& \textbf{73.2\%} & 73.6\%\\
\bottomrule
\end{tabular}
\end{table}

\begin{table}[htbp]
\caption{The classification results with less training samples on Pubmed}
\label{tab-lesspubmed}
\centering
\begin{tabular}{c|ccc}
\toprule
\multirow{2}{*}{Method} & \multicolumn{3}{c}{Label Rate} \\
                        & 0.03\%   & 0.05\%   & 0.1\%    \\
\midrule
LP\cite{lp}& 58.3\%   & 61.3\%   & 63.8\%   \\
GCN\cite{gcn}& 51.1\%   & 58.0\%   & 67.5\%   \\
Co-training\cite{citation_1}& 55.5\%   & 61.6\%   & 67.8\%   \\
Self-training\cite{citation_1}& 56.3\%   & 63.6\%   & 70.0\%   \\
Union\cite{citation_1}& 57.2\%   & 64.3\%   & 70.0\%   \\
Intersection\cite{citation_1}& 55.0\%   & 58.2\%   & 67.0\%   \\
MultiStage\cite{multistage}& 57.4\%   & 64.3\%   & 70.2\%   \\
M3S\cite{multistage}& 59.2\%   & 64.4\%   & 70.6\%   \\
GWNN\cite{gwnn}& 71.7\%   &    70.4\%      &       72.9\%   \\
GCNII\cite{gcnii}& 71.8\%   & 71.7\%   & 73.4\%   \\
\midrule
\textbf{DeepGWC}&\textbf{73.4\%}   &  \textbf{74.1\%}      & \textbf{75.3\%} \\
\bottomrule
\end{tabular}
\end{table}

\begin{figure*}[hptb]
    \centering
    \subfigure[Cora]{
    \includegraphics[width=5cm]{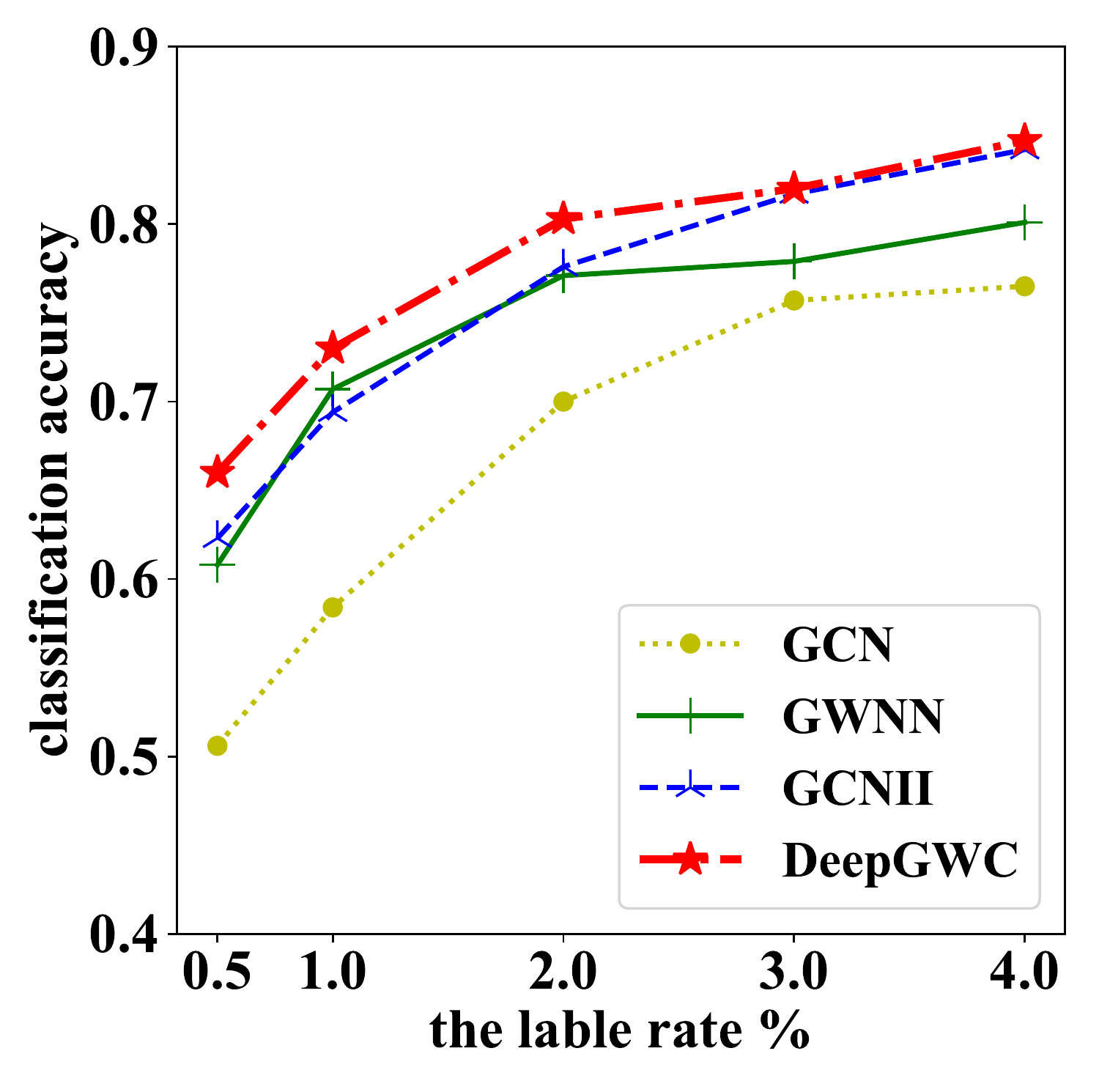}
    }
    \subfigure[Citeseer]{
    \includegraphics[width=5cm]{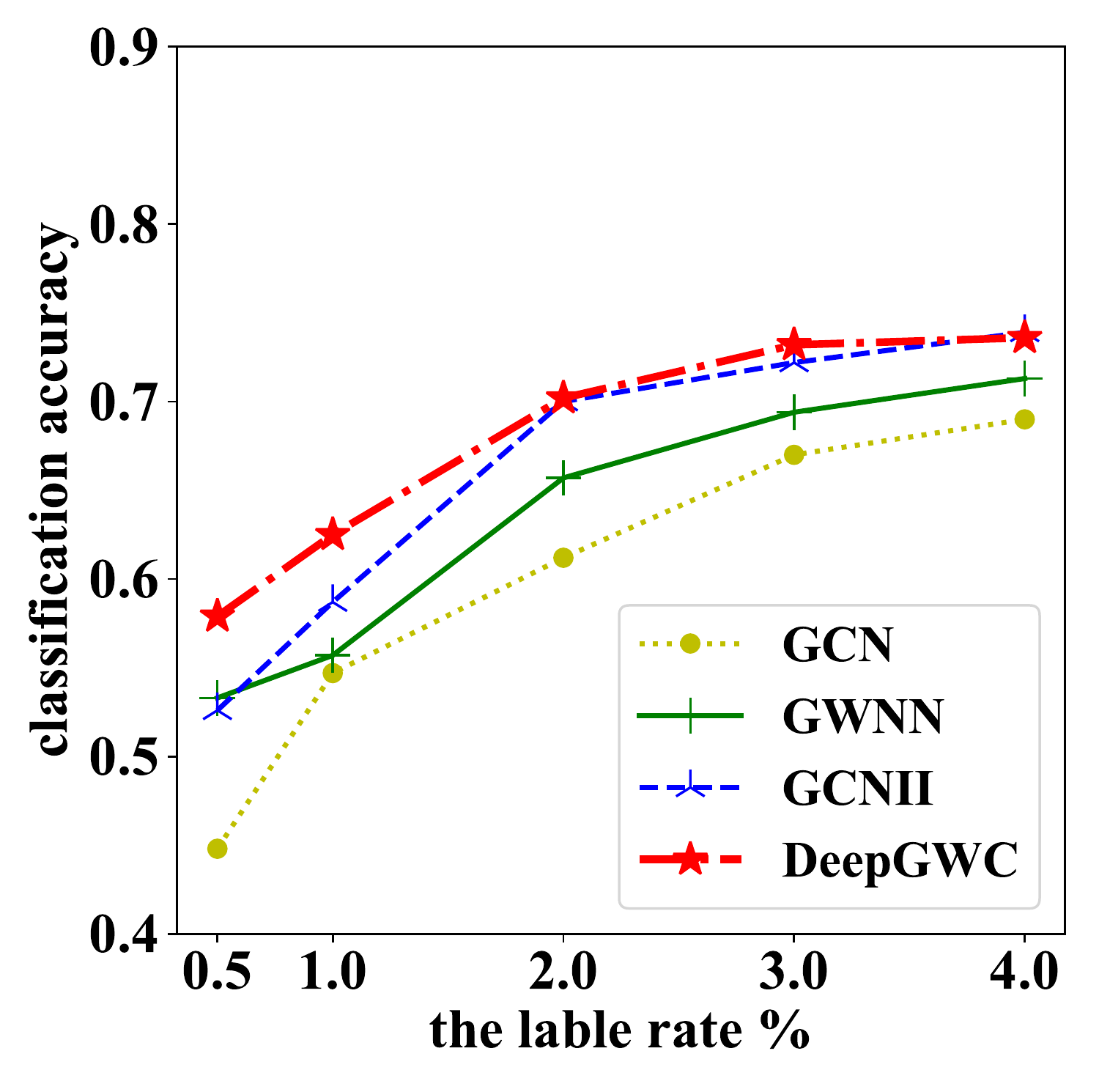}
    }
    \subfigure[Pubmed]{
    \includegraphics[width=5cm]{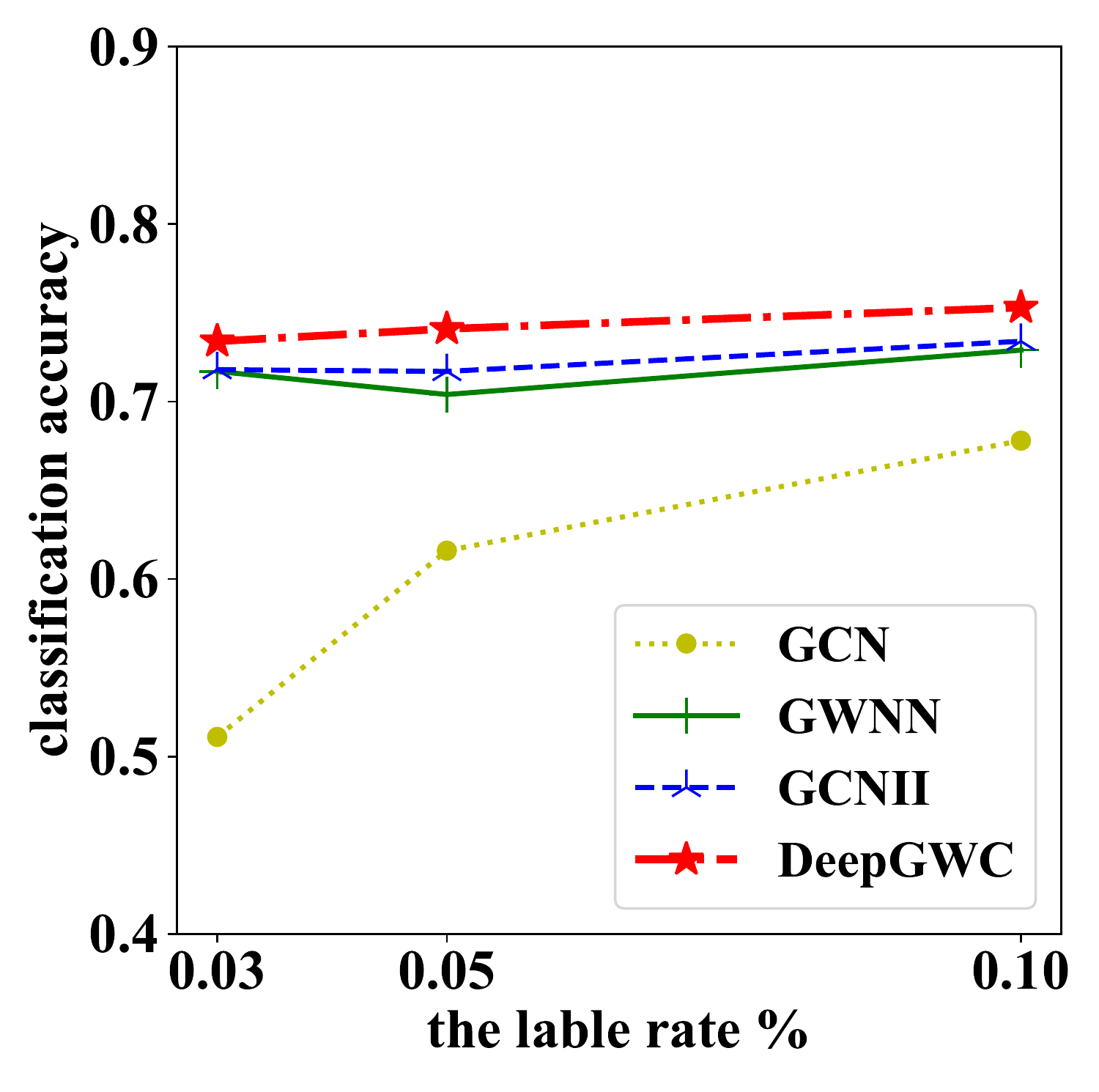}
    }
    \caption{The classification accuracy with less training samples.}
    \label{fig-labelrate}
\end{figure*}

We compare DeepGWC with GCN\cite{gcn}, GWNN\cite{gwnn}, and GCNII\cite{gcnii} in the case of less training data more carefully. The comparison results are shown in Fig. \ref{fig-labelrate}. It should be noted that DeepGWC performs better when there is less training data. In other words, the gap between DeepGWC and GCNII\cite{gcnii} is more obvious in the case of lower label rates. In experiments on Cora and Citeseer with the label rates of 0.5\% and 1\%, DeepGWC exceeds GCNII\cite{gcnii} by 4.1\% on average while the gap is 0.73\% for the label rates of 2\%, 3\%, and 4\%. Besides, comparing GCN\cite{gcn} and GWNN\cite{gwnn} that are both shallow models, we find that it is also in the case of lower label rates that GWNN\cite{gwnn} yields better results than that with normal label rates. The comparison between DeepGWC and GCNII\cite{gcnii} and the comparison between GWNN\cite{gwnn} and GCN\cite{gcn} with fewer training nodes further prove the excellent capability of feature extraction of graph wavelets. It suggests that graph wavelets succeed in extracting useful information even with less labeled nodes.

\section{Conclusion}
In this paper, we present a deep graph wavelet convolutional network called DeepGWC. We modify the filtering matrix of graph wavelet convolution to make it adaptable to deep graph convolutional models. Applying the combination of Fourier bases and wavelet ones, the proposed DeepGWC model with the reuse of residual connection and identity achieves better performance than existing deep graph models. Extensive experiments on semi-supervised node classifications are conducted. On Cora, Citeseer, and Pubmed, the experimental results demonstrate that our DeepGWC model outperforms the baselines with 2, 4, 8, 16, 32, and 64 layers and yields new state-of-the-art results.

\section*{Acknowledgment}

This work was supported in part by the National Key R\&D Program of China under Grant No.2017YFB1302200 and 2018YFB1600804, by TOYOTA TTAD 2020-08, by a grant from the Institute Guo Qiang (2019GQG0002), Tsinghua University, by Alibaba Group through Alibaba Innovative Research Program, by Tencent Group, and by ZheJiang Program in Innovation, Entrepreneurship and Leadership Team (2018R01017).

% \begin{comment}

% \end{comment}
% \bibliographystyle{IEEEtran}
% \bibliography{mybib}

\end{document}